# A JOINT APPROACH TOWARDS DATA-DRIVEN VIRTUAL TESTING FOR AUTOMATED DRIVING
## THE AVEAS PROJECT[1]

**Leon Eisemann**[6,12], **Mirjam Fehling-Kaschek**[1], **Silke Forkert**[2], **Andreas Forster**[3], **Henrik Gommel**[4], **Susanne Günther**[1], **Stephan Hammer**[5], **David Hermann**[6,11], **Marvin Klemp**[7], **Benjamin Lickert**[1], **Florian Lüttner**[1], **Robin Moss**[1], **Nicole Neis**[6,7], **Maria Pohle**[1], **Dominik Schreiber**[1], **Cathrina Sowa**[8], **Daniel Stadler**[1], **Janina Stompe**[9]*, **Michael Strobelt**[6], **David Unger**[10], **Jens Ziehn**[1]

[1] *Fraunhofer EMI / IOSB / IVI,* [2] *PTV Planung Transport Verkehr GmbH,* [3] *Continental Automotive Technologies GmbH,* [4] *GOTECH GmbH,* [5] *Spiegel Institut Ingolstadt GmbH,* [6] *Porsche Engineering Group GmbH,* [7] *Karlsruhe Institute of Technology KIT,* [8] *dSPACE GmbH,* [9] *UnderstandAI GmbH,* [10] *Allianz Zentrum für Technik,* [11] *Technical University of Munich TUM,* [12] *Stuttgart Media University HdM*
*\*Contact address: Karlsruhe, Baden-Wuerttemberg, 76185, Germany*
*Corresponding author's phone: +49 173 6718809, e-mail: janina.stompe@understand.ai*

**ABSTRACT**: With growing complexity and responsibility of automated driving functions in road traffic and growing scope of their operational design domains, there is increasing demand for covering significant parts of development, validation, and verification via virtual environments and simulation models.

If, however, simulations are meant not only to augment real-world experiments, but to replace them, quantitative approaches are required that measure to what degree and under which preconditions simulation models adequately represent reality, and thus allow their usage for virtual testing of driving functions. Especially in research and development areas related to the safety impacts of the "open world", there is a significant shortage of real-world data to parametrize and/or validate simulations – especially with respect to the behavior of human traffic participants, whom automated vehicles will meet in mixed traffic.

This paper presents the intermediate results of the German AVEAS research project (www.aveas.org) which aims at developing methods and metrics for the harmonized, systematic, and scalable acquisition of real-world data for virtual verification and validation of advanced driver assistance systems and automated driving, and establishing an online database following the FAIR principles.

## 1. Introduction

Given the increasing demand for covering significant parts of development, validation, and verification of future automated driving functions (SAE Level 3+) via virtual environments and simulation models, as outlined in the abstract, the role of simulations must change from tools answering particular questions under manually-defined parameters, towards tools determining general system properties under "open world" conditions with only broad specifications, for example concerning the operational design domain (ODD). These properties must be determined quantitatively rather than just qualitatively in face of the perspective that these automated driving (AD) functions will not be distinguished into "perfectly safe" and "not perfectly safe", but rather into "acceptably low residual risk" and "inacceptable residual risk". At the same time, failure modes of automated vehicles will become more complex with the growing complexity of systems and ODDs [1], such that even the manual design of simulation models may likely limit the breadth of factors that can be covered in virtual tests.

This motivates the goal of increasingly founding simulation models (both their parametrization and, in some cases, their design) on real-world data obtained by systematic acquisitions not tailored for a specific "system under test" (SUT, i.e., the automated vehicle function in question), but rather developing a "digital twin" of the real word capable of finding system limitations that were, in some cases, not even known to human experts beforehand.

As potential sources for such data-driven simulations, various databases and formats already exist that represent real-world traffic data and data for simulation in safety testing. One of the most comprehensive efforts available in simulatable accident data today is GIDAS (German In-Depth Accident Study) and the TASC database[2]. However, GIDAS is based on accident reconstruction data, which are limited to the vehicles immediately involved in the accidents, contains significant uncertainty regarding driving behavior prior to the accident and is limited to a provision of only 2,000 accident reconstructions per year. TASC contains reconstructed accidents with a basic approach to create synthetic scenarios but also only accounts for the vehicles directly involved in the accident. Other existing sources include, e.g., the highD, SHRP 2 NDS and NGSIM datasets, which, however, have a very specific perspective on driving behavior aside of accidents and with limited scalability to comprehensive validation.

---

[1] This publication was written in the context of the AVEAS research project (www.aveas.org), funded by the German Federal Ministry for Economic Affairs and Climate Action (BMWK) within the program "New Vehicle and System Technologies".

[2] https://www.vufo.de/tasc/

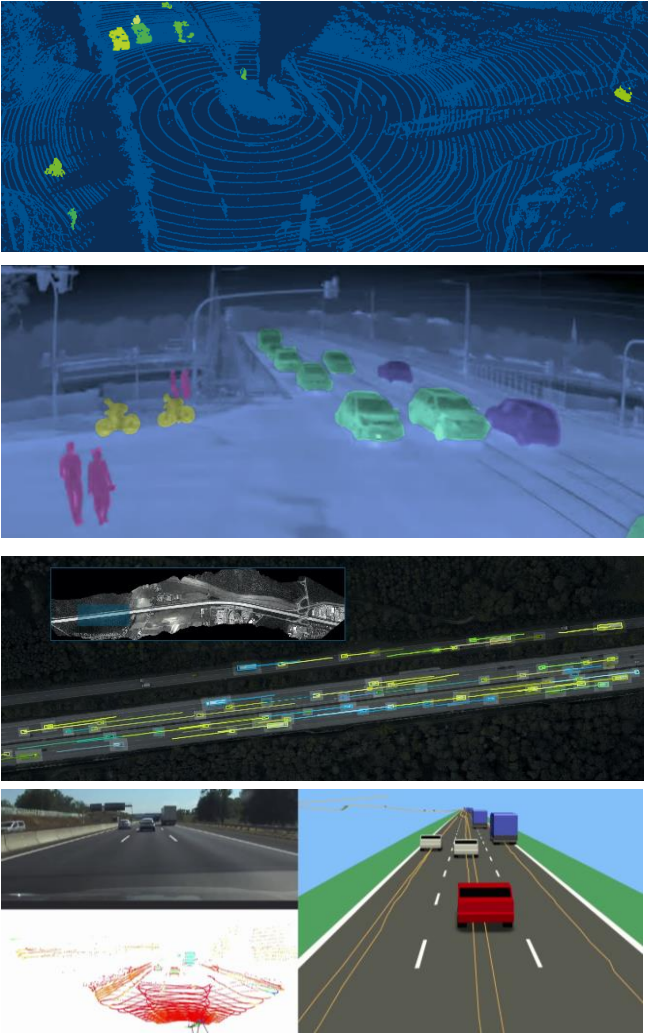

Fig. 1: Different acquisition methods used in AVEAS (cf. also [2]); from top to bottom: LiDAR point cloud recorded at an intersection with detected dynamic objects; infrared image of an urban intersection with detected road user types; aerial image of an interstate highway from ultralight aircraft with detected vehicles and motion trajectories; vehicle based data acquisition with camera, LiDAR and RADAR sensors, as described in [3] (with extracted roads and vehicles shown in *esmini*).

The AVEAS research project (en.: "Acquisition, Analysis and Simulation of Traffic Situations for AD Safety Assurance"), funded by the German Federal Ministry for Economic Affairs and Climate Action, comprises a consortium from industry, standardization, and research in advanced driver assistance systems (ADAS), AD and safety. The project sets out to develop methods and metrics for the systematic and scalable acquisition of real-world data for virtual ADAS/AD verification and validation, and to establish an online database following the FAIR principles [2], with an approach outlined in Fig. 5.

In this paper, we present intermediate project results on risk definition, harmonization aspects concerning different acquisition methods, and data specifications to create a harmonized scenario database, which shall support virtual verification and validation methods. Finally, perspectives of data-driven virtual testing on practical use cases in AD are discussed.

## 2. Harmonization Needs

The project aims to cover real-world traffic data from urban, rural, and highway driving situations. These are acquired at accident hotspots with specific homogeneous accidents (and thus likely places for repeated behavior under similar risky conditions) via infrastructure-, road vehicle-, and aircraft-based sensors, complemented by driving behavior studies. This combination enables the coverage of use cases ranging from the continuous monitoring of hazardous intersections up to following interstate highway traffic at the same speed over extended time periods and spatial coverage from both ground and aerial perspectives (see [2]).

This, however, creates a need to harmonize the provided data, such that approach-specific relevant information is preserved, while common minimal standards across all approaches enable a collective processing for a sufficiently wide range of use cases. To allow a unified data processing and evaluation, a common data structure and attribute specification across all acquisition methods (see Fig. 1), and subsequent method-specific automated processing pipelines must be implemented.

Furthermore, the surveyed traffic data not only cover different areas (each with specific traffic conditions and common situations), but also cover situations with a varying degree of criticality and thus show the need to find common measures for distinction.

## 3. Fields of Risk Harmonization Needs

For the scope of data-driven virtual testing, we adopt a notion of "risk" for the vehicle system that follows the concept of Safety of the Intended Functionality (SOTIF) specified in ISO 21448, relating to the correlated safety contributions of the overall system as specified, and the dynamic environment within the ODD, instead of risks arising from failures of individual components in the sense of functional safety (ISO 26262). Such risks have received particular attention in the context of the 2016 crash of a Tesla Model S and a tractor semitrailer in Florida [1], which was the result of a wide range of contributing factors from automated perception over system design, active and passive safety, to human factors and human–machine interface design, cumulating in the overall accident scenario depicted (as an approximation) in Fig. 2.

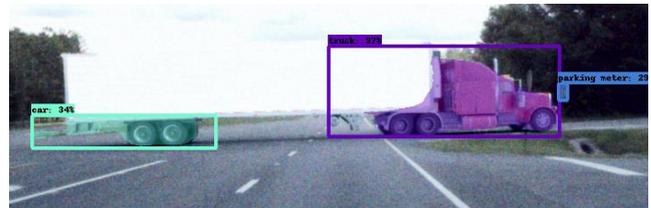

Fig. 2: Qualitative resimulation of the 2016 accident in which a Tesla Model S ran under a semitrailer [1] – possible only through the coincidence of the irregular driving behavior of the human truck driver, limitations of the Tesla's sensors and sensor fusion principles, and insufficient interactions between the human passenger with the Tesla system. Data-driven virtual testing allows to not only find such coincidences, but also to quantify the risk by probability and impact. Exemplary ML object detection results use a Mask R-CNN trained on MS COCO.

We hence propose that situational criticality of an automated driving function in early forms of mixed traffic is governed by a wide range of factors, especially time and space and environmental condition factors. In this perspective, "risk" is a complex, multi-parametric effect that requires rather accurate quantification to distinguish high-exposure risks from unrealistic or highly unlikely scenarios, leading to the demand of providing data of stochastically significant volume and accuracy.

With a focus on the harmonization of risk measures, this, again, requires harmonized data structures and attribute specifications of providable data, across acquisition methods and risk perspectives, bridging different acquisition methods as well as scenario contexts, from accident reconstruction data up to the observation of irregular behaviors in normal traffic. Thus, time–space-related metrics need to be able to present criticality differences not only between accidents with severe and minor injury probabilities but between normal driving situations, incidents, and accidents as well. Furthermore, metrics shall not only allow to assess situations in whole, but also at discrete time points. Metrics for measuring risk and different sources of risks shall be introduced in the following.

### 3.1 Measuring time–space-related risk

Conventional time–space related risk measures use varying dynamical parameters such as time, distances, speeds, accelerations, or jerks mostly in relation to other actors in situations. In general, they are distinguishable by their:
- provision of threshold values,
- ability to assess time-discrete risk (as opposed to only an overall risk throughout the situation),
- applicability to normal driving situations (as opposed to only accident situations),
- use of multiple counterfactual simulations to account for dynamical accident avoidability, and
- applicability to various situations (as opposed to only certain situations).

Risk measures were preferred for the harmonized data format which especially show a high applicability for various situations. This, on the one hand, provides a comparability between different situations, and on the other hand minimizes the amount of criticality measures which may only help to assess a limited number of situations. At the current state of AVEAS, gTTC [4] and PrET [5] are considered as common risk measures and are complemented by basic measures such as time and distance headway. Risk measures using counterfactual simulation to account for alternative situational outcomes are not yet implemented but may be added after the setup of the scenario database, which provides the basis for sampling methods to create synthetic alternative situations.

### 3.2 Risk relating to human behavior

Risks relating to human behavior refer to risks arising from human errors and/or behavior outside of the norm or of traffic laws during full human control of a vehicle, which impact the safety of nearby automated vehicles. To integrate this type of behavior into a validation process, behavior models within the state of the art are evaluated and parametrized by optimization methods based on the aforementioned data, and new data-driven model approaches are explored that are capable of extracting critical human behaviors (as, for example, seen in human traffic rule violations in the case of the Tesla accident [6]) for use in simulations of novel scenarios.

As part of the model parametrization and simulation framework in AVEAS, we use PTV Vissim [7], which incorporates the Wiedemann99 model [8] to simulate the following behavior of the traffic participants. Further, a modified version of the Sparmann model [9] accounts for the lane change dynamics of the agents. Additional parameters, like the distribution of desired velocities or the distribution of maximal and desired acceleration/deceleration, can be set. To optimize the underlying model parameters, we use an iterative procedure. Therefore, we calculate the log-likelihood of the distributions of observables of interest, such as the vehicle velocities for recorded data given a simulated estimation. Using a Nelder–Mead algorithm [10], the parameters are then varied, such that the likelihood to observe the recorded data within the simulated data is maximized.

As reported in [2], an initial parameter optimization based on an aerial test data set was done. Here, we optimized the distributions of the desired velocity for two agent classes, namely cars and trucks. Since we assumed Gaussian distributions, this led to four parameters to be optimized, i.e., two means and two standard deviations. The overall results validate the feasibility of our optimization procedure.

### 3.3 Risk relating to automated driving functions

This category summarizes risks relating to automated driving functions in full automated mode, meaning the sole responsibility for the dynamic driving task (DDT) lies with the AD function entirely over the considered time interval. This introduces novel types of risks caused, e.g., by limitations in sensors, perception, and planning algorithms as presented in Fig. 1 and Fig. 3. To model these effects, real-world data collections for typical automated vehicle sensors are conducted specifically under adverse conditions.

For different sensor modalities, different sensor-specific characteristics are considered. In the case of camera sensors, we evaluate phenomena such as limited dynamic range or the rapid adaption of gain and white balance [11]. In the automotive context, these are mostly caused by the roadside environment, namely gantries, bridges, or tunnels. Furthermore, different weather conditions as well as times of day are considered to record the impact on perception. In the case of LiDAR sensors, the AVEAS project particularly considers highly reflecting environments, e.g., parking lots and wet roads. These are of particular interest as these can lead to multipath reflections causing erroneous distance estimations or even missing detections.

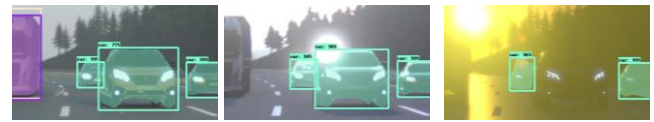

Fig. 3: Specific parameter variations with data-driven models enable the sensitivity analysis of ML methods, as shown on the example of Mask R-CNN w.r.t. lighting conditions (simulated in OCTANE[3], cf. Fig. 1).

### 3.4 Risk relating to the transition between automated and manual driving modes

Risks relating to the transition between automated and manual driving modes include handovers by the driving function as well as active interventions by the human driver into the automated system. Human–machine interaction studies are conducted in a virtual reality (VR) based driving simulator to evaluate risk factors in challenging transition scenarios. To achieve a high level of immersion, the interior of the SUT is modeled accurately in a VR environment along with a corresponding physical hardware dummy mounted on a 6 DOF (degrees of freedom) motion system driving the simulator platform.

This setup allows the simulation and monitoring of driving situations with focus on drivers' behavior for varying human–machine interfaces (HMI) existing in the SUT as well as novel interfaces regarding their respective layout and content.

Test cases include specific handover scenarios between human drivers and automated driving functions in various variations. All data including the exterior scenario and vehicle-internal parameters such as driver pose, motions, eye tracking information, and control inputs, are recorded to be systematically included in the database.

## 4. Harmonized Data Format

### 4.1 General aspects

Related standards for virtual testing are the ASAM standards OpenDRIVE for map data, OpenSCENARIO for logical scenario variation and sampling, and OpenLABEL for the annotation of sensor and scenario data. We employ a combination of these standards and propose a specification for instantiating OpenLABEL for map-referenced trajectories as a common format abstraction for highly diverse traffic data acquisition methods.

---

[3] www.octane.org

The database shall give information and filter criteria for the assessment of safety- and assistance-related systems, as well as for traffic flow behavior, by providing static and dynamic data on traffic and accident situations. Thereby, the relation between the different contents shall be preserved.

Each acquisition method (cf. also [2] for an overview) offers specific advantages with regard to detectable information for specific properties (e.g., vehicle light states, driver state) which may not be detectable by other methods but may also have restrictions (e.g., unobservable road areas by occlusion or limited sensor field of view – Fig. 4). Therefore, the data format should be able to include optional information. This leads to high standardization, harmonization, and expansion needs of used labels and formats within the database and a comprehensive structuring of information.

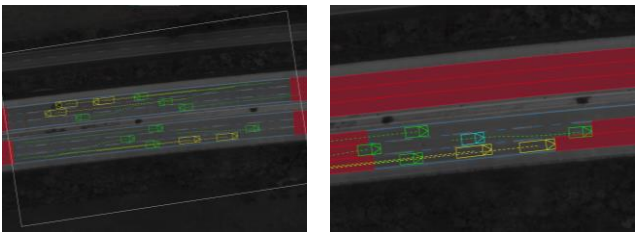

Fig. 4: Unobservable road areas are annotated consistently, e.g., for aerial data (left, unobservability due to limited camera frame field of view) and vehicle-recorded data (right, limited primarily by sensor range and occlusion by ground objects).

4.2 Harmonized data format structure

ASAM OpenLABEL is a JSON-based format that allows an easy machine parsing and hierarchical structuring of data and was developed to standardize the annotation and labeling both of multi-sensor data streams and scenario files. OpenLABEL only sets basic requirements for units and formats and the naming of objects. In general, these objects can be divided into "Labelling", "Tagging", and "Metadata" objects on a high level. Thereby, the level of labeling data generally represents data with spatiotemporal changes, whereas objects on the level of tagging data and metadata represent static information. Each object can contain further nested information [12]. OpenLABEL is, therefore, highly customizable, and able to meet the beforementioned requirements.

In the following, some of the content and its allocation in the developed AVEAS-specific OpenLABEL structure shall be presented. How the specific requirements can be met is best illustrated on the level of labeling data within the structured OpenLABEL format (see Tab. 1). Generally, on this level, information needs to be distinguished by the object it is related to and, therefore, special distinct identifiers and hierarchical sub-objects are introduced. The objects "Context", "Objects", "Events", and "Frames" can be distinguished on the level of "Labeling". The objects "Context" and "Objects" describe measures that are static during the whole scenario, respectively, for the actors in the scenario. The object "Events" describes predefined found events in the scenario. Via previously defined objects (ParticipantID), nested information on the movement of all involved objects in the event can be implemented. The object "Frames" allows to especially specify spatio-temporal changes. Since time–space related risk measures are often indicated between two objects, a further sub-object is introduced to denote risk measures of one ParticipantID to multiple further ParticipantIDs.

From Tab. 1, the ability of the created format to link risk measures to different objects they relate to can further be seen: Risks relating to human behavior are linked to a certain ParticipantID at a specific time, conventional time–space related risks are linked to one ParticipantID in relation to another ParticipantID at a specific time, whereas risks relating to the transition between automated and manual driving modes may be linked to special EventIDs and risks relating to automated driving functions are referred to special weather conditions linked to a ScenarioID.

Further information was assigned to the level of "Metadata". This includes information on objects, such as "coordinate systems" (and their relation to each other), references to associated "resources" (such as OpenDRIVE files) and references to used "ontologies". Considering the topic of ontologies within the OpenLABEL format, two aspects shall be especially mentioned. For each information (not only on the level of "Metadata") a distinct OntologyID may optionally be referred to. OntologyIDs are further introduced within the object "Ontologies", which enlists the used ontologies. However, focusing on the development of a scenario database, this leads to high storage requirements if ontology information is recurrently provided for each scenario. In the context of a scenario database, it may therefore seem more appropriate to disregard the possibility of quoting and referring to ontologies as long as ontologies remain unchanged over time and are common for all acquisition methods.

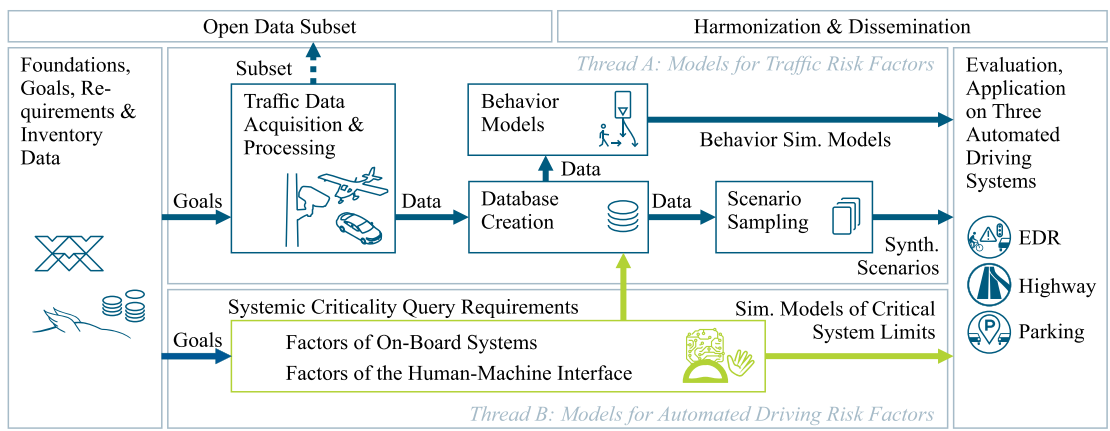

Fig. 5: Overview of the project approach. In Thread A, behavior data is acquired in public traffic, processed to yield map-referenced trajectories stored in a central scenario database. This data is used both to parametrize and evaluate behavior models and to sample relevant yet realistic scenarios following plausible distributions according to the acquired data. In Thread B, novel risks are analyzed introduced by characteristics of automated driving functions that are not – or only scarcely – seen in current public road traffic, namely technical factors of on-board systems while the automated function is exclusively responsible for the dynamic driving task (DDT), and transitions between automated and manual control as factors of the human–machine interface (HMI). The obtained methods for virtual testing and validation are applied to three automated driving systems.

Tab. 1: Information on the level of "Label" data (optional information is marked by *)

| Object 1/ ID 1 | Sub-object 2/ ID 2 | Sub-object 3/ ID 3 | Content |
|---|---|---|---|
| "Context"/ ScenarioID | – | – | - General conditions (weather, lighting, traffic, road surface) |
| "Objects"/ ParticipantID | – | – | - road user type, <br> - dimensions, <br> - speed range, <br> - dynamics during collision*, <br> - steering wheel positions* |
| "Events"/ EventID | – | – | - type of event*, <br> - associated time interval |
|  | "Objects"/ ParticipantID | – | - classification of movement during event* |
| "Frames"/ FrameID | – | – | - timestamp |
|  | "Objects"/ ParticipantID | – | - 3D bounding box, <br> - state of different lights (indicator, brake light, etc.)*, <br> - position (w.r.t. road lane coordinates and to world coordinates), <br> - speed limit*, <br> - traffic condition*, <br> - dynamic (speed, acceleration*, yaw*, pitch* roll*), <br> - risk measures of human behavior |
|  | "Objects"/ ParticipantID | | - time–space related risk measures with reference to other objects |

Normally, OpenLABEL distinguishes between the objects "Metadata" and "Tagging". "Tagging" describes data which helps to describe a scenario in general (and not for isolated, geometric, or spatiotemporal information as the objects "Frames" or "Events") and which includes keywords for searching and filtering scenarios within a database. In contrast, the object "Metadata" shall be used for informative purposes, like file versions, authorships, etc. However, to minimize the number of objects, no distinction was made between metadata and tagging data in the AVEAS OpenLABEL format, since both information refer to the scenario in general. Tab. 2 summarizes the content for the objects on the level of metadata.

**5. Simulation Setup and Outlook**

After collecting and harmonizing the data recorded by the different acquisition methods, the database can be used to parameterize data-driven microscopic traffic models. These models can then be used to create simulations of critical situations, in which the performance of AD functions under risky conditions can be assessed.

As previously mentioned, the simulation framework in the AVEAS project consists of PTV Vissim for the simulation of traffic behavior, the Carla and OCTANE simulation frameworks for automated vehicle simulation, and the VR driving simulator operated by GOTECH. An overview of models in the state of the art is given in [13]. These frameworks and models are extended over the duration of the project to increasingly support data-driven models based on the acquired data.

Tab. 2: Information on the level of "Metadata" (optional information is marked by *)

| Object/ RelationID 1 | Sub-object/ RelationID 2 | Content |
|---|---|---|
| "Coordinate Systems"/ ScenarioID | – | - EPSG of used world coordinate system, <br> - longitude/latitude (in world coordinates) of local coordinate system |
|  | Coordinate SystemID | - type of coordinate system (static, local, sensor), <br> - children and parent coordinate systems |
| "Metadata"/ ScenarioID | – | - creation time, <br> - acquisition method (stationary LiDAR, stationary infrared, aerial RGB video, vehicle-based sensors (LiDAR, camera, radar)), <br> - data use restrictions, <br> - origin (reconstructed, sampled, original), <br> - area (urban, highway, rural), <br> - scenario duration, <br> - range of dynamic values (speed, distances, etc.) |
| "Resources"/ ScenarioID | – | - path to OpenDRIVE file |
| "Ontologies"/ OntologyID | – | - URI to ontology definitions*, <br> - boundaries* (if only a subset of ontologies shall be used) |

In upcoming steps, we plan to extend the data basis of the model parametrization for human behavior in traffic (in particular, the *longitudinal behavior*, i.e., the choice of speeds, accelerations, and distances given the traffic scenario), described in Sec. 3.2. Having determined an optimized traffic model parameter set, we can then create arbitrary variations of realistic traffic to sample critical situations – thus, enabling us to search the database for specific scenarios of interest. One such scenario could be a critical lane change. The recorded traffic situation (number of vehicles, relative positions, velocities) is used as a starting point for the simulations. Subsequently, relevant parameters are then varied with the range of found distributions for this specific scenario type. For the given example, this would include the distance of the lane changing car to the approaching faster vehicle on the other lane and using an agent-based traffic model to simulate the effect of the parameter change on the criticality of the situation. The approaching vehicle would be forced to decelerate more strongly once the distance to the lane changing car is reduced, which increases the criticality of the situation. Once the deceleration needed to circumvent a collision exceeds the maximal deceleration of the vehicle model, we will observe a crash, i.e., the limiting case of critical situations.

The outlined procedure could potentially be used in the future to assess automated driving functions by simply assigning its function characteristics to one or several of the simulated vehicles. In

Copyright © 2023 Society of Automotive Engineers of Japan, Inc. All rights reserved

this way it could be possible to check if the driving function causes critical situations either directly, or as induced accidents, e.g., strong decelerations of agents needed to avoid collisions with the automated vehicle or within the surrounding traffic.

Corresponding work to parametrize the *lateral behavior* of vehicles inside their own lane by the recorded data is given in [14]. The model there proposed a two-level approach based on a state-discrete Markov model and state-continuous Brownian noise model that can be "trained" on recorded lateral behavior data and enables potential extensions into more comprehensive environment-dependent models once sufficient data is available.

The generated traffic scenarios will then be used in Carla, CarMaker, and OCTANE to simulate automated driving functions (adding sensor, dynamics, and bus / interface simulations) and an event data recording system, in PCCrash to reconstruct vehicle dynamics during collisions as well as in the VR driving simulator studies to, again, refine behavior models of humans based on interior monitoring.

## 6. Conclusion

In this paper, we have discussed the requirements for a joint approach to systematically acquire real-world data for use in data-driven simulations of AD functions – with the perspective that such simulations are key to enabling the efficient safety validation of AD functions if and only if they prove able to quantitatively predict safety performance of a system under "open world" conditions.

This approach requires a systematic and harmonized approach of acquiring real-world data, processing it in a highly automated way to provide annotations, metadata and fusing data from different sources, such as exterior traffic data, vehicle interior data, sensor / perception data, and environmental conditions relevant for AD safety evaluation. Establishing fundamental concepts for this joint approach is the goal of the AVEAS research project, with a particular scope of harmonizing risk measures and data formats from fundamentally different data acquisition methods and providing a FAIR dataset of critical traffic scenarios from various perspectives.

The intermediate project results presented here indicate that the implementation of a highly automated data acquisition and utilization pipeline is a feasible and scalable approach to achieve data-driven models for the safety validation of AD functions, with a suitable method-independent data format based on the ASAM OpenLABEL standard defined as a basis.

Thereby, the joint and systematic acquisition of data, a common structured data format and ontology is intended to resolve key questions in quantitative safety estimation – above all, enabling to prioritize system limits not just by their criticality level but also according to their statistically expected exposure.

In this context, the results of the AVEAS project are intended as an initial feasibility study and as a specification of necessary commonly usable standards across different domains and tasks. Unlocking the potentials of the approach and such data for safety gains in automated driving and road traffic development, however, will require continuous and international efforts.